\documentclass{article}

\usepackage{arxiv}

\usepackage[utf8]{inputenc} 
\usepackage[T1]{fontenc}    
\usepackage{hyperref}       
\usepackage{url}            
\usepackage{booktabs}       
\usepackage{amsfonts}       
\usepackage{nicefrac}       
\usepackage{lipsum}
\usepackage{graphicx}
\usepackage{multirow}

\title{Hybrid ANN-SNN Pipeline with Local Plasticity}

\author{
Denis Larionov \\
  Chuvash State University, \\
  Cheboksary, Russia \\
  LLC NForm, Moscow, Russia \\
  \texttt{denis.larionov@gmail.com} \\
  \And
Khairutin Shtanchaev \\
  Dagestan State Technical University, \\
  Makhachkala, Russia \\
  LLC 1T, Moscow, Russia \\
  \And
Mikhail Kiselev \\
  Chuvash State University, \\
  Cheboksary, Russia \\
  LLC NForm, Moscow, Russia \\
  \And
Mikhail Korovin \\
  LLC 1T,  Moscow, Russia \\
  \And
Ivan Tugoy \\
  LLC 1T, Moscow, Russia \\
  LLC NForm, Moscow, Russia \\
}

\begin{document}

\maketitle

\begin{abstract}
This work proposes a hybrid ANN-SNN pipeline that effectively leverages the rich embeddings of pretrained artificial neural networks (ANNs) to enable high-performance spiking neural networks (SNNs). The architecture couples a pretrained EfficientNet encoder with a CoLaNET spiking classifier. We convert the encoder's activations into spike trains via rate-coding and train the subsequent SNN classifier using local, biologically inspired learning rules, bypassing end-to-end gradient propagation. This approach achieves 99.09\% accuracy on a 64-class ImageNet benchmark, demonstrating performance on par with conventional deep networks. The work presents a biologically plausible and efficient framework for adapting powerful pretrained encoders to downstream spiking neural network tasks.
\end{abstract}


\section{Introduction}
\label{sec:intro}

Spiking Neural Networks (SNNs) are a class of neural networks based on temporal and event-driven signal transmission, inspired by biological neural systems \cite{Maass1997}. Unlike traditional Artificial Neural Networks (ANNs), which process information using continuous-valued activations, SNNs operate with discrete events (spikes), enabling more efficient and biologically plausible computation \cite{Gerstner2002}.

In contrast to ANNs, which use matrix multiplications (suitable for GPUs), SNNs are fully asynchronous, use event-driven computations, and have several other features that make GPUs an inefficient hardware platform for their execution. However, when implemented on specialized neuromorphic hardware, SNNs may offer high energy efficiency and low latency \cite{Ivanov2022}. Since spikes are generated only when necessary, SNN-based hardware systems can significantly reduce computational power consumption, making them suitable for embedded systems, robotics, and edge devices with limited resources \cite{Roy2019}.


There are several approaches to training SNNs:
\begin{itemize}
\item ANN is converted into SNN by transforming activations into spike rates or spike times. This approach is known as ANN2SNN \cite{Rybka2026}.
\item Gradient-based methods. Training is performed in the SNN domain using backpropagation through time (BPTT) \cite{Neftci2019}.
\item Local learning. The synaptic weight updates depend only on the activity and state of the neurons connected by that synapse \cite{Kiselev2020a}.
\end{itemize}

Although gradient-based or conversion approaches often achieve better accuracy, local plasticity rules are the primary candidates for building energy-efficient, low-latency systems capable of continuous learning \cite{Larionov2024a}. However, SNNs with local rules are very difficult to use in complex architectures. Designing SNNs remains a craft-like art. One way to mitigate this problem is through hybridization.

Hybrid architectures combining ANN and SNN models aim to exploit the strengths of both paradigms. While ANNs provide powerful representation learning, SNNs offer event-driven processing, temporal dynamic, and local learning rules \cite{Pei2019}.

This work proposes a hybrid ANN-SNN architecture that couples a pretrained EfficientNet encoder \cite{Tan2019a} with a CoLaNET spiking classifier \cite{Kiselev2024a}. We convert the encoder's activations into spike trains via rate-coding and train the subsequent SNN classifier using local, biologically inspired learning rules, bypassing end-to-end gradient propagation. This approach achieves 99.09\% accuracy on a 64-class ImageNet benchmark, demonstrating performance on par with conventional deep networks. The source code and experimental results are open-source and available at: https://gitflic.ru/project/dlarionov/hybridsnn.

\section{Related works}
\label{sec:related_works}

Works \cite{Kiselev2024a, Kiselev2024b, Kiselev2025b} introduce the CoLaNET (Columnar Layered Network) architecture for classification tasks. CoLaNET is a single-layer fully connected network, so we cannot directly use it for large image classification except for small-sized pictures. For instance, \cite{Kiselev2024b} used the MNIST dataset, interpreting pixel intensity values as spike frequencies. To process larger images, the authors propose using dimensionality reduction methods or encoders.

Work \cite{Larionov2025b} studies the CoLaNET architecture in a continual learning protocol, again on simple datasets (MNIST, EMNIST, PermutedMNIST). This study shows that the network’s resistance to catastrophic forgetting critically depends on the generality of feature representations across tasks. To address the generality problem, the authors propose adding a hierarchically structured encoder that converts large images into semantically rich feature representations but with low generality.

This paper continues the above works by proposing to couple a spiking neural network classifier (CoLaNET) with a hierarchically structured encoder. Such encoders are widely available in convolutional neural networks (CNNs).

A popular encoder choice for image classification in deep learning is EfficientNet. Work \cite{Tan2019a} shows that a pretrained EfficientNet can serve as a fixed feature extractor for classification on new datasets. \cite{Kolesnikov2019} shows that with a frozen encoder, fine-tuning on new data requires only a few training examples. \cite{Tan2019a} uses EfficientNet as a frozen or partially frozen backbone for detection tasks.

Hybridizing ANN and SNN architectures looks promising for hardware implementation. For example, the Tianjic project \cite{Pei2019} builds neuromorphic processors capable of efficiently running large hybrid networks in inference mode by reusing the same circuit blocks for different network types (ANN and SNN). Another example that combines a frozen convolutional encoder with a plastic SNN-based classifier is the Akida platform. \cite{Ivanov2022} provides examples of using Akida for one-shot learning in face recognition.

\section{Background}

\subsection{CoLaNET architecture}
\label{sec:colanet}

CoLaNET (Columnar Layered Network) architecture \cite{Kiselev2024a, Kiselev2024b, Kiselev2025b, Kiselev2025c} embodies columnar organization in SNNs with local learning, neuronal competition, modulated plasticity, and gating mechanisms. Designed for supervised classification, CoLaNET's key feature integrates anti-Hebbian plasticity (degrading weights) with modulated plasticity (strengthening weights), counteracting degradation and enabling effective learning.

CoLaNET consists of multiple identical modules (columns), with column count matching class count. Each column has trainable fully connected input layer (many L neurons), intermediate processing layer (single BIASGATE neuron), and output layer (single OUT neuron). L neurons inside one column correspond to significantly distinctive instances (sub-classes) of one class. All neurons are described by the simplest LIF (leaky integrate-and-fire) model with slight modifications described later.

\begin{figure}[htbp]
  \centering  
  \includegraphics[width=0.9\textwidth]{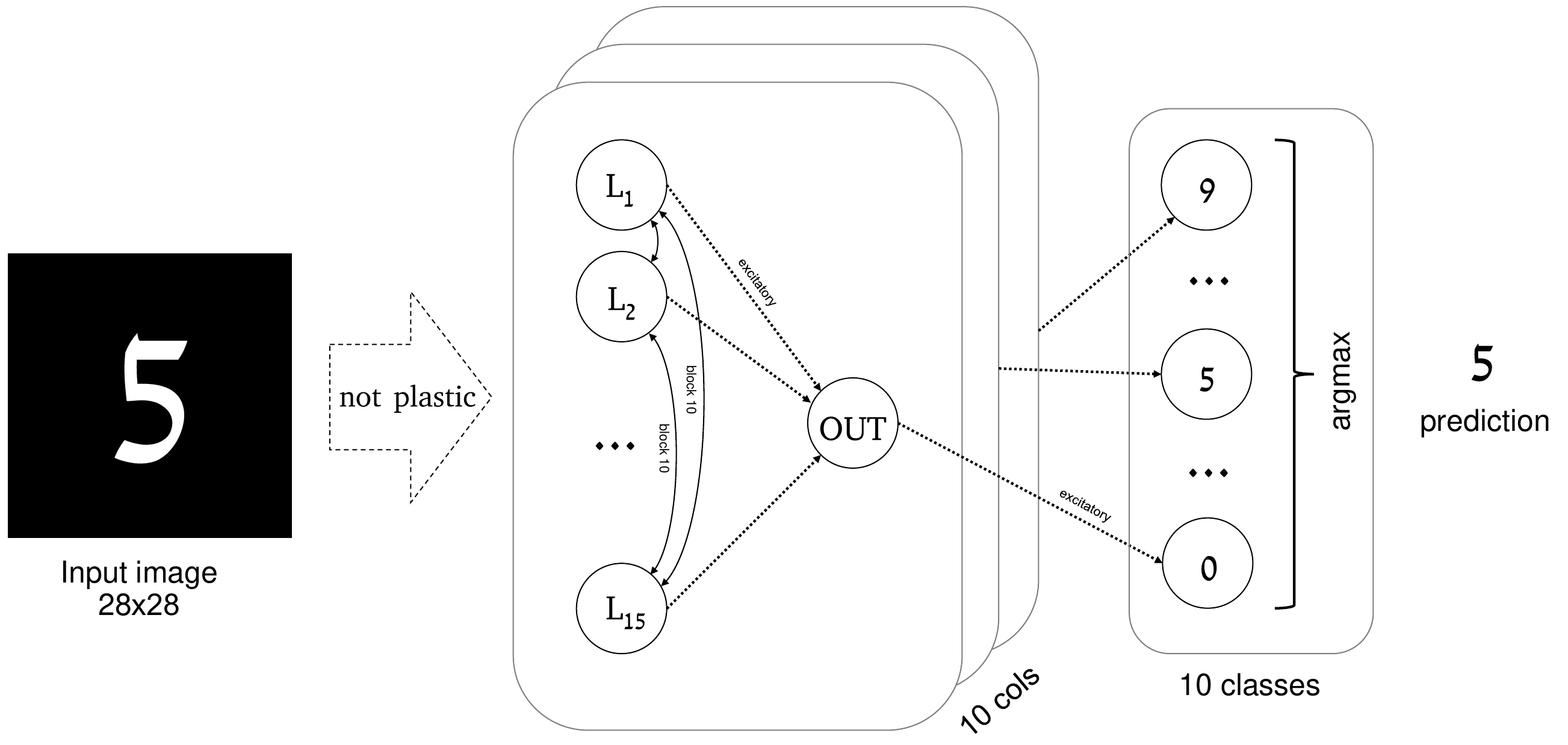}
  \caption{CoLaNET architecture, inference regime. The image is presented over 10 time steps, followed by 5 time steps of silence (empty image). The neuron L that first generates a spike suppresses other L neurons within the same column and propagates the spike to the OUT neuron. The class is determined by the column producing the highest number of spikes.}
  \label{fig:inference}
\end{figure}

CoLaNET converts pixel intensities to spike sequences using rate coding, where spike generation probability at each time step equals pixel intensity ratio (0 intensity = zero probability, 255 = maximum probability). Images appear for 10 time steps, followed by 5 silence steps.

In the inference regime (Figure \ref{fig:inference}), some of the trained L neurons fire in response to the presentation of an object belonging to the corresponding class. It forces the correct OUT neuron to fire (it is treated as a classification decision). The final prediction is obtained by majority vote across an ensemble of several independently trained CoLaNET instances.

\begin{figure}[htbp]
  \centering  
  \includegraphics[width=0.9\textwidth]{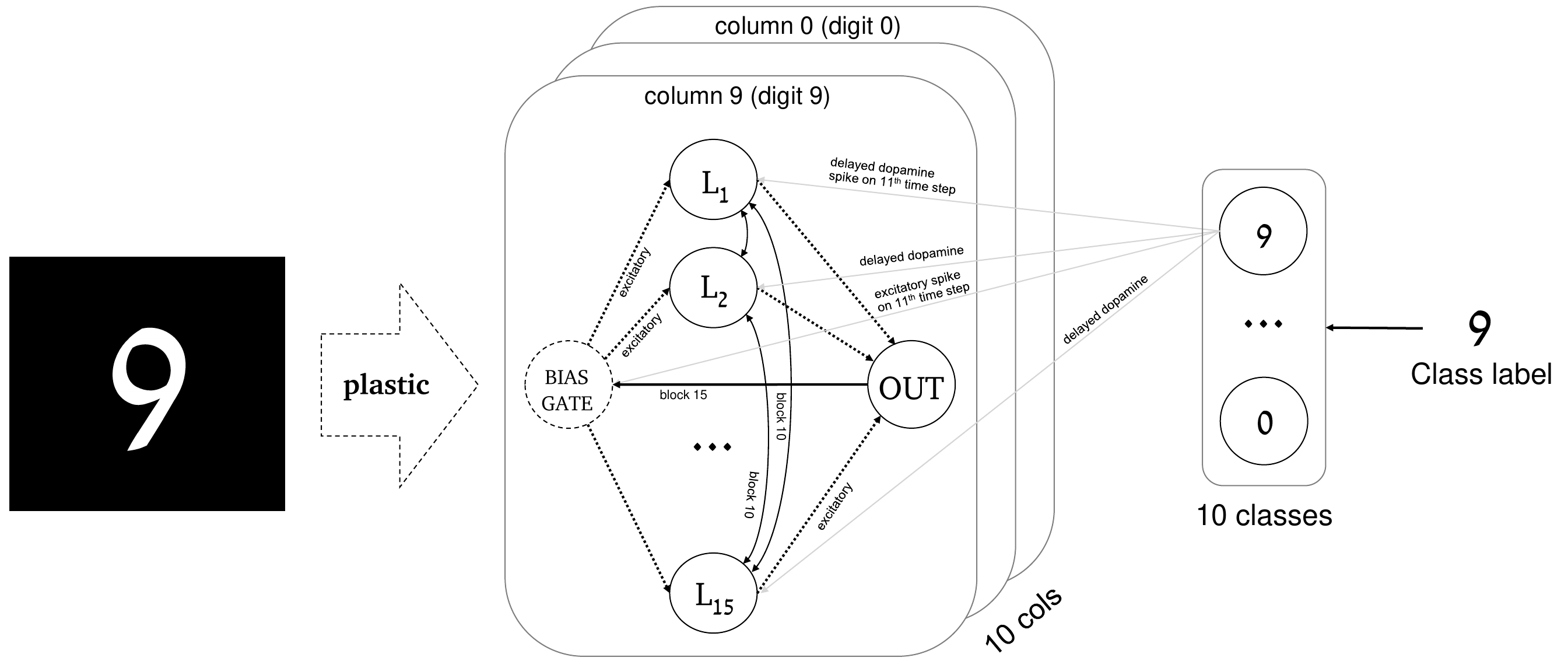}
  \caption{CoLaNET architecture, learning regime. Learning is governed by three factors: anti-Hebbian plasticity (weakening synaptic weights), dopamine-modulated plasticity (strengthening weights), and periodic synaptic renormalization. The label is presented as a single spike at 11th time step. Each class stimulates its own column. On the first step, a reset is performed, which clears all set blocks.}
  \label{fig:training}
\end{figure}

CoLaNET training relies on neuronal competition for activation when presented with input patterns. In the learning regime, the network obtains spikes encoding the current image as well as information about class (label) of this image. The latter has the form of a single spike emitted on the 11th time step by the input node corresponding to the current class label (Figure \ref{fig:training} right).

Initially, all plastic synapses of L‑neurons are zero, so input cannot fire them. Learning starts when a class label node sends a spike to the BIASGATE neuron at the 11th time step, causing BIASGATE to fire and strongly excite all L‑neurons in its column. Due to internal stochastic noise, their membrane potentials differ; the WTA mechanism selects the neuron with the highest potential as the winner. The same class label node also sends a dopamine spike directly to all L‑neurons. This triggers dopamine plasticity: synapses that received spikes shortly before the neuron fired are potentiated because the dopamine spike arrives shortly after firing. Consequently, one L‑neuron slightly strengthens connections to recently active inputs. On subsequent similar stimuli, this neuron has a higher chance to win again, further potentiating the same synapses. To ensure different L‑neurons learn different instances of the same class, CoLaNET uses (1) competition via constant total synaptic weight per neuron (strengthening some synapses weakens others uniformly) and (2) threshold variation proportional to the sum of positive plastic weights. After enough repetitions, some L‑neurons fire from input alone, activate the OUT neuron, and block BIASGATE for that object presentation.

\subsection{EfficientNet encoder}
\label{sec:eNet}

The EfficientNet family \cite{Tan2019a} provides a principled approach to scaling CNNs. Unlike traditional CNNs that independently increase depth, width, or resolution, EfficientNet uses a compound scaling method that balances all three dimensions simultaneously.

Models range from B0 (lightest) to B7 (most accurate). This scaling strategy yields strong performance: EfficientNets consistently achieve higher accuracy with fewer parameters and FLOPs than many CNN architectures \cite{Tan2019a}.

We select EfficientNet-B3, pretrained on ImageNet \cite{deng2009imagenet}, as the feature extraction backbone. We use weights from the PyTorch repository (see source code for details). B3 contains approximately 10.8M parameters and requires 1.8B FLOPs per image in inference mode. ImageNet is a large‑scale visual recognition benchmark containing 1000 object classes, 1,281,167 training images, 50,000 validation images, and 100,000 test images. Due to its scale and diversity, it has become a standard foundation for training and evaluating general‑purpose visual representations.

We extract intermediate representations from the output of the global average pooling layer, yielding a compact 1536-dimensional feature vector per input image (Figure \ref{fig:B3model}). These vectors encode high-level semantic information learned during ImageNet pretraining. They are then fed into the downstream hybrid ANN-SNN classification pipeline, which separates convolutional feature extraction from spike-based processing.

No additional augmentations, such as rotation, flipping, or color modification, were employed.

\section{Methods}

\subsection{Dataset subsets}
\label{sec:benchmarks}

In experiments, researchers often use subsets of the original ImageNet instead of the full dataset. These subsets not only contain fewer images per class but also have all images resized to a consistent dimension. The original dataset requires approximately 150 GB of storage and is not publicly accessible, so this work utilizes one of the popular subsets available on the Kaggle platform (see source code for details), where all images are resized to 256×256×3 and centered.

Moreover, training large SNNs poses considerable challenges due to hardware resource constraints; therefore, we additionally restrict the number of classes. We construct two balanced subsets that focus on experimental evaluation while preserving the generality of the original categories. Each subset is built from visually distinct object classes, ensuring a wide range of semantic concepts and appearance variations (illumination, background, scale, viewpoint).

The first subset comprises 26 categories, with approximately 80 images per class, totaling 15,691 images. The second subset contains 64 categories, with roughly 90 images per class, totaling 35,179 images. In both subsets, the data are split into non‑overlapping training and testing sets using an 80/20 ratio and are shuffled.

The complete set of selected ImageNet classes for the 26-class subset includes: \texttt{airliner}, \texttt{banana}, \texttt{bicycle\_built\_for\_two}, \texttt{bookshop}, \texttt{bow}, \texttt{crayfish}, \texttt{crib}, \texttt{fire\_engine}, \texttt{fountain}, \texttt{granny\_smith}, \texttt{lakeside}, \texttt{laptop}, \texttt{microwave}, \texttt{park\_bench}, \texttt{parking\_meter}, \texttt{persian\_cat}, \texttt{pineapple}, \texttt{pizza}, \texttt{running\_shoe}, \texttt{siberian\_husky}, \texttt{sports\_car}, \texttt{strawberry}, \texttt{street\_sign}, \texttt{traffic\_light}, \texttt{violin}, and \texttt{volcano}.

The complete set of selected ImageNet classes for experiments with 64 classes includes: \texttt{abaya}, \texttt{academic\_gown}, \texttt{accordion}, \texttt{acorn}, \texttt{acorn\_squash}, \texttt{admiral}, \texttt{affenpinscher}, \texttt{african\_chameleon}, \texttt{african\_elephant}, \texttt{african\_hunting\_dog}, \texttt{agaric}, \texttt{airedale}, \texttt{airliner}, \texttt{albatross}, \texttt{alligator\_lizard}, \texttt{altar}, \texttt{american\_black\_bear}, \texttt{american\_chameleon}, \texttt{american\_egret}, \texttt{analog\_clock}, \texttt{angora}, \texttt{ant}, \texttt{apron}, \texttt{arabian\_camel}, \texttt{arctic\_fox}, \texttt{armadillo}, \texttt{artichoke}, \texttt{ashcan}, \texttt{australian\_terrier}, \texttt{axolotl}, \texttt{baboon}, \texttt{badger}, \texttt{bakery}, \texttt{ballpoint}, \texttt{banana}, \texttt{band\_aid}, \texttt{barbell}, \texttt{barber\_chair}, \texttt{barn}, \texttt{barn\_spider}, \texttt{barracouta}, \texttt{barrel}, \texttt{barrow}, \texttt{baseball}, \texttt{basenji}, \texttt{basset}, \texttt{bassoon}, \texttt{bathtub}, \texttt{beacon}, \texttt{bearskin}, \texttt{beaver}, \texttt{bedlington\_terrier}, \texttt{bee\_eater}, \texttt{beer\_glass}, \texttt{bell\_cote}, \texttt{bell\_pepper}, \texttt{bicycle\_built\_for\_two}, \texttt{bighorn}, \texttt{binder}, \texttt{birdhouse}, \texttt{bison}, \texttt{bittern}, \texttt{black\_and\_gold\_garden\_spider}, and \texttt{black\_grouse}.

\subsection{Hybrid architecture}
\label{sec:architecture}

\begin{figure}[!ht]
    \centering
    \includegraphics[width=1\linewidth]{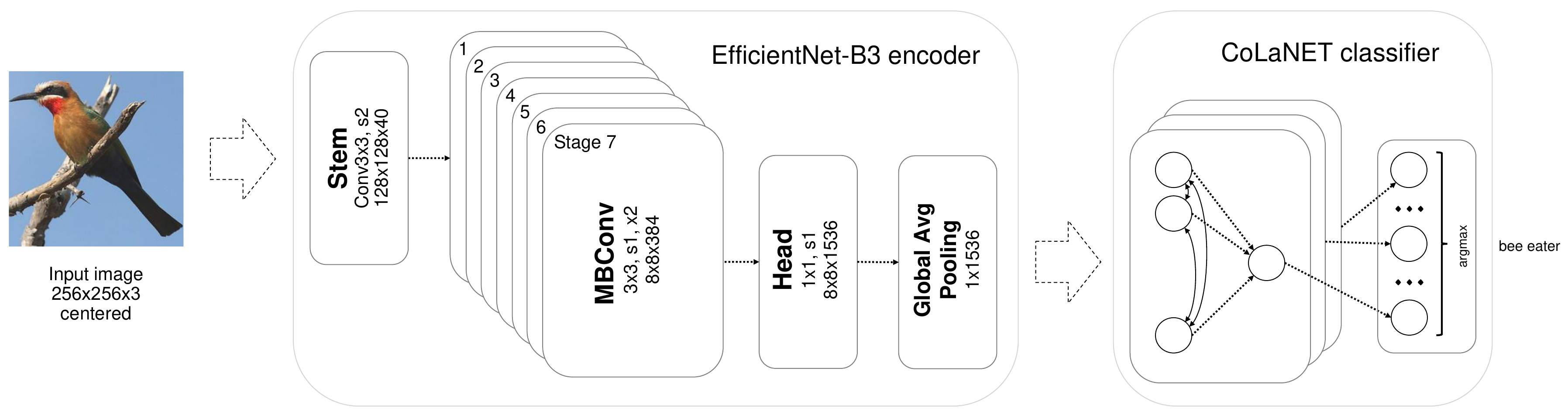}
    \caption{The architecture consists of an EfficientNet‑B3 encoder followed by a CoLaNET classifier. The input 256×256×3 image first passes through a stem 3×3 stride‑2 convolution, yielding 128×128×40 feature maps, which are then processed by a series of MBConv layers, progressively reducing spatial size to 8×8 with 384 channels. A head 1×1 convolution expands the depth to 1536 channels, followed by global average pooling to produce a 1536‑dimensional vector. This vector is rate‑coded by CoLaNET over 10 presentation steps with a subsequent 5‑step silence period, and finally classified via argmax (e.g., into classes like "bee eater")}
    \label{fig:B3model}
\end{figure}

Figure~\ref{fig:B3model} presents the proposed hybrid ANN-SNN architecture. It combines convolutional feature extraction via EfficientNet-B3 with spike-based processing in CoLaNET.

The conversion from activation vectors to spike trains proceeds in three steps. First, we suppress all negative activation values to zero—conceptually equivalent to ReLU. Second, we use an expertly chosen threshold. Activations at or above this threshold produce spikes at every timestep. Third, we linearly scale sub-threshold activations to a spike count between 0 and 10. Higher activations yield more spikes.

This encoding preserves relative activation magnitudes while producing a temporally distributed spike representation suitable for CoLaNET.


\section{Experiments}
\label{sec:experiments}

We implemented all training and evaluation experiments using the ArNI-X framework \cite{ArNI-X}. Since CoLaNET was originally designed in ArNI-X, the framework includes all necessary mechanics.

Each training sample is presented to the network exactly once — no epoch-based repetition is employed. Learning is thus online and single-pass: the network adapts its synaptic weights continuously as new examples arrive, without revisiting previously seen data.

First, we evaluated the hybrid architecture on the 26-class subset. The model achieved stable learning and high classification performance, exceeding 99\% test accuracy. Encouraged by this result, we next tested the larger 64-class subset. This setting introduces greater semantic diversity and classification complexity, allowing us to assess scalability and robustness under more challenging conditions. This work omits the details of the 26-class experiment, but they are available in the source code.

\subsection{Network optimization}
\label{sec:opt}

Training SNNs introduces additional complexity in hyperparameter tuning, as the optimization landscape is highly non-uniform and generally non-smooth. A common approach under such conditions is to apply a genetic algorithm (GA) followed by gradient-based or stochastic descent from the best configuration \cite{Kiselev2026}.

In this work, we adopted the optimization scheme and baseline configuration from \cite{Kiselev2024b}, where CoLaNET was optimized for MNIST handwritten digit recognition. We used a similar hyperparameter set, adjusted for the architectural evolution of CoLaNET, for the 64-column version of the network. We also fixed the number of CoLaNETs in the ensemble at 15.

We optimized the hyperparameters using a genetic algorithm. The accuracy metric reached 99.09\% at the 11th generation and remained unchanged after the 12th generation. Each generation comprised 300 candidates, with each candidate encoding 9 parameters that controlled synaptic plasticity, weight dynamics, stochastic stimulation, and network structure. At each generation, we evaluated all candidates on the classification task and ranked them by test accuracy (fitness). We preserved the best individuals via elitism and generated the remaining population through crossover between selected parents.

Each experiment consisted of 28,143 training and 7,036 testing presentations. Each presentation exposed the image to the spiking network for 15 timesteps. Thus, training required 422,145 timesteps, testing required 105,540 timesteps, and a full single-network evaluation totaled 527,685 timesteps.

The GA optimized 9 hyperparameters: the amplitude of positive synaptic updates (\texttt{DWPOS}), the ratio of negative to positive weight changes (\texttt{RELDWNEG}), the stability resource adaptation rate (\texttt{STABILITYRESOURCECHANGERATIO}), the maximum and minimum synaptic weights (\texttt{MAXWEIGHT} and \texttt{MINWEIGHT}), the number of learning groups (\texttt{NLEARNINGGROUPS}), the excess-weight threshold regulation (\texttt{WTHREXCESS}), the number of silent synapses (\texttt{NSILENTSYNAPSES}), and the saturation coefficient (\texttt{SATURATION}).

After GA optimization, we fine-tuned the best configuration using stochastic descent (SD). This step typically yields a modest improvement in accuracy; however, in our case, it did not produce any gain, as the metric remained essentially flat.

In total, approximately 3,794 full experiments were executed during the optimization campaign. The complete optimization process took approximately three days on a GPU cluster equipped with two NVIDIA GeForce RTX 4090 cards (24 GB memory each). The software environment used CUDA Driver 13.0 and Runtime 12.8.

\subsection{Results}
\label{sec:results}

The best classification accuracy optimization reached is 99.09\%. Table~\ref{tab:ga_parameters} lists the optimal parameter values found.

\begin{table}[!ht]
\centering
\caption{Best hyperparameter configuration obtained by optimization}
\label{tab:ga_parameters}
\begin{tabular}{|l|c|}
\hline
\textbf{Parameter} & \textbf{Optimized value} \\
\hline
DWPOS                           & 0.0986409      \\
RELDWNEG                        & 0.504279665      \\
STABILITYRESOURCECHANGERATIO    & 0.0644199    \\
MAXWEIGHT                       & 1.18684      \\
MINWEIGHT                       & 1.65234      \\
NLEARNINGGROUPS                 & 1          \\
WTHREXCESS                      & 0.0019573    \\
NSILENTSYNAPSES                 & 1222        \\
SATURATION                      & 2.07474        \\
\hline
\end{tabular}
\end{table}

\begin{figure}[!ht]
    \centering
    \includegraphics[width=0.7\linewidth]{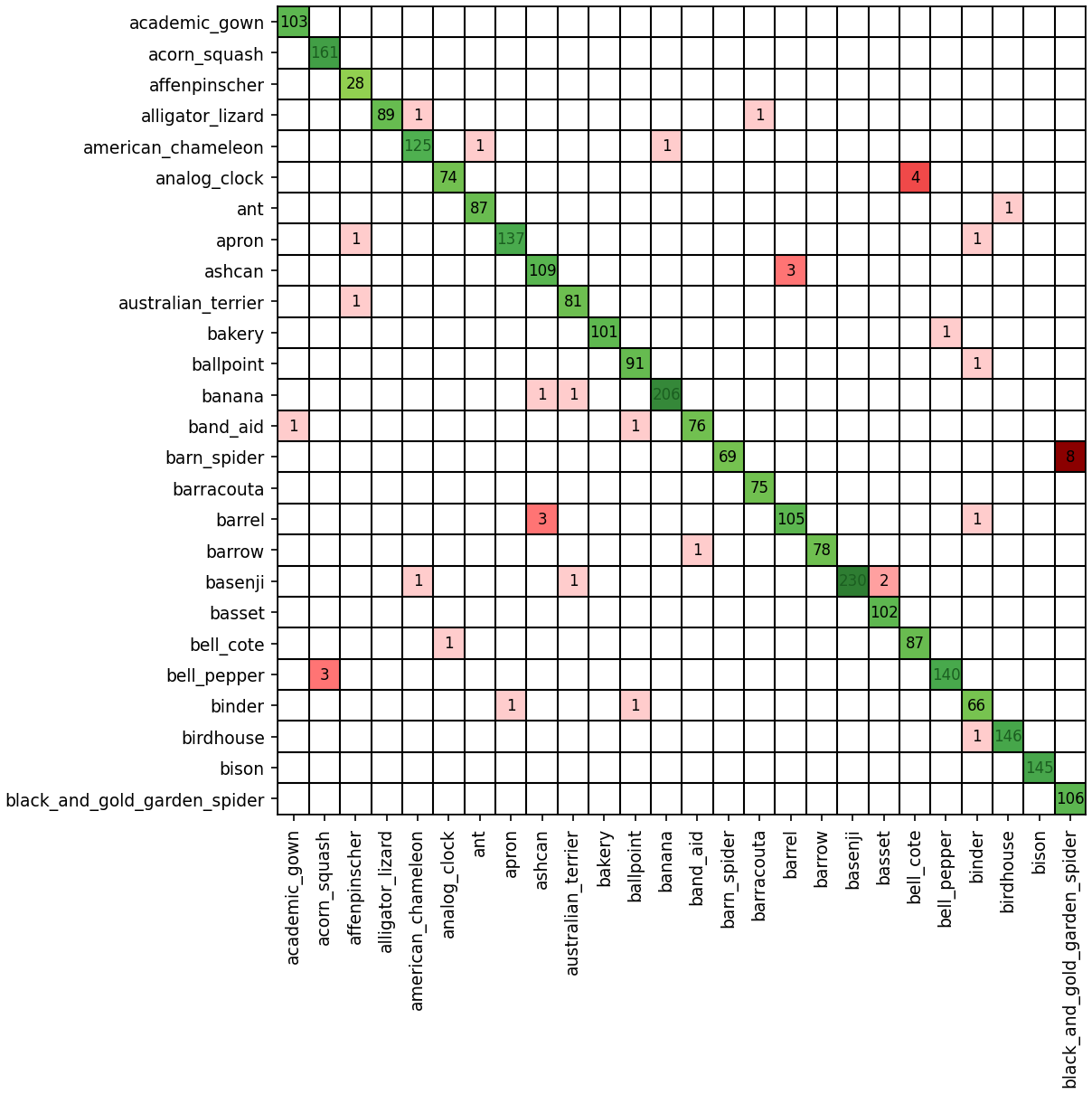}
    \caption{Confusion matrix for classes with more than one prediction error.}
    \label{fig:сonfusion_matrix}
\end{figure}

The Figure~\ref{fig:сonfusion_matrix} shows the confusion matrix for classes with more than one prediction error. Approximately half of the 64 classes achieve zero prediction errors. The largest confusion, with 8 errors, occurs between \textit{barn\_spider} and \textit{black\_and\_gold\_garden\_spider}—different spider species that are difficult even for humans to distinguish. The second largest confusion, with 4 errors, involves \textit{analog\_clock} and \textit{bell\_cote}, as bell tower images frequently contain analog clocks. Three errors occur between \textit{ashcan} and \textit{barrel}, reflecting the visual similarity of some trash cans to barrels. No other significant misclassifications appear.

For comparison, we trained a single-layer conventional ANN on the extracted feature representations. The SNN optimization produced a configuration with 1 neuron per class, so 64 neurons per CoLaNET times 15 networks in the ensemble yields approximately 1,000 plastic neurons. An ANN with 1,000 neurons, without any additional tools, achieves 98\% accuracy after the first epoch and reaches 99.25\% by the 12th epoch. We provide the implementation details in the accompanying source code. Thus, the hybrid architecture achieves classification accuracy comparable to ANN.

\section{Discussion}
\label{sec:discussion}

EfficientNet-B3 encoder was pretrained on the full ImageNet-1K dataset, from which our 64-class subset was drawn. The encoder was therefore exposed to the test images during its original training, and the reported accuracy may partly reflect retrieval of previously learned representations rather than learning of entirely novel categories. We acknowledge this openly while emphasizing that our primary objective here is different: we aim to demonstrate that a frozen ANN encoder can be effectively coupled with a spiking classifier initialized with zero weights and trained exclusively with local plasticity rules. Whether the encoder generalizes to genuinely unseen distributions is a separate question. A rigorous test would require a dataset with classes absent from the encoder's pretraining — a direction we consider important for future work.

Notably, the reported accuracy is achieved in a single-pass, online learning regime: each training example is presented only once, and the network never revisits previously seen data. This is a non-trivial property that distinguishes local plasticity approaches from gradient-based methods, which typically require multiple epochs of shuffled replay.

Using an ensemble rather than a single CoLaNET classifier increases the parameter count, but the cost is minimal. All instances share one frozen encoder, so expensive feature extraction runs once per image. Prior works \cite{Kiselev2024a, Kiselev2024b, Kiselev2026} show that CoLaNET ensembles yield the largest gain in accuracy precisely when individual classifiers are weak. One costly encoder, many cheap spiking decoders — this asymmetric architecture prioritizes accuracy while keeping the decision stage efficient and biologically plausible.

\section{Conclusion}
\label{sec:conclusion}

We proposed a hybrid ANN-SNN architecture that couples a frozen EfficientNet-B3 encoder with a CoLaNET spiking classifier, trained exclusively with local, biologically inspired plasticity rules. On a 64-class ImageNet subset, this pipeline achieved 99.09\% classification accuracy, matching the performance of conventional deep networks.

This design embodies a pragmatic and powerful paradigm: decouple feature extraction from decision-making to let each paradigm (ANN and SNN) operate where it excels. The ANN encoder efficiently distills high-dimensional images from the real world into compact, semantically rich feature vectors — a task that remains very difficult for end-to-end SNN training. The SNN classifier then processes these representations using event-driven computation and local learning, opening a direct path to ultra-low-power neuromorphic hardware without sacrificing accuracy.

Even in its current hybrid form, implementing this architecture on neuromorphic hardware is justifiable for on-device continual learning — following the blueprint established by projects like Akida \cite{Ivanov2022}, where a frozen convolutional encoder feeds a plastic on-chip spiking classifier. However, from an energy-efficiency standpoint, the heavy ANN encoder remains a bottleneck, consuming the bulk of computational resources. The most promising direction for future work, therefore, is converting the ANN encoder into a fully spiking model that operates on event-based principles and continues to be frozen. Such a conversion would eliminate the activation-to-spike encoding step entirely and unlock the full potential of neuromorphic processors, delivering end-to-end event-driven computations with dramatically lower power consumption and latency.

By combining the representational power of pretrained deep networks with the efficiency and adaptability of local spiking computation, hybrid ANN-SNN architectures chart a practical route toward autonomous, low-power, and continuously learning intelligent systems. The transition from hybrid to fully spiking pipelines stands as the natural next milestone on this path.

\section{Conflict of interest}
This is a preprint of the Work accepted for publication in Optical Memory and Neural Networks (Information Optics), copyright 2026, https://pleiades.online.

\bibliographystyle{unsrt}  
\bibliography{main}

\end{document}